%% file: _paper.tex
\documentclass[sigconf,nonacm]{acmart}

\AtBeginDocument{%
  \providecommand\BibTeX{{%
    \normalfont B\kern-0.5em{\scshape i\kern-0.25em b}\kern-0.8em\TeX}}}

\copyrightyear{2019}
\acmYear{2019}
\setcopyright{rightsretained}

\acmConference{~~}{~~}{~~}
\acmDOI{~~}
\acmISBN{~~}
\acmBooktitle{~~}

\usepackage{caption}
\usepackage{subcaption}
\usepackage{multirow}

\usepackage{enumitem} 
\usepackage{dblfloatfix}

\begin{document}

\title{Privacy Attacks on Network Embeddings}


\author{
Michael Ellers$^1$, Michael Cochez$^2$, Tobias Schumacher$^1$, Markus Strohmaier$^{1,3}$, Florian Lemmerich$^1$}
\affiliation{%
  \institution{\textsuperscript{1}RWTH Aachen University \textsuperscript{2}VU Amsterdam \textsuperscript{3}GESIS - Leibniz Institute for the Social Sciences}
}


\renewcommand{\shortauthors}{M. Ellers et al.}

\def\G#{G}
\newcommand{\Gw}[1]{\G{}_{/\set{\v{#1}}}}
\def\Gprime#{\G{}'}

\newcommand{\Gprimew}[1]{\Gprime{}_{/\set{\v{#1}}}}
\def\Gpp#{\G{}''}
\def\Gstar#{\G{}^{*}}
\def\Gstarprime#{\G{}^{*'}}
\newcommand{\g}[1]{g_{#1}}

\def\V#{V}
\newcommand{\Vw}[1]{\V{}_{/\set{\v{#1}}}}
\def\Vprime#{\V{}'}
\def\Vpp#{\V{}''}
\renewcommand{\v}[1]{v_{#1}}
\def\Vstarprime#{\V{}^{*'}}

\def\E#{E}
\def\Eprime#{\E{}'}
\def\Epp#{\E{}''}
\newcommand{\e}[1]{e_{#1}}

\def\GVE#{\G(\V{},\E{})}
\def\GVEprime#{\Gprime(\Vprime,\Eprime)}
\def\GVEpp#{\Gpp(\Vpp,\Epp)}

\def\EMB#{\mathcal{E}}
\newcommand{\EMBw}[1]{\EMB{}_{/\set{\v{#1}}}}
\newcommand{\Emb}[1]{\EMB{}(#1)}
\newcommand{\emb}[1]{\vec{u_{#1}}}
\newcommand{\Embw}[2]{\Emb{#1}_{/\set{\v{#2}}} }
\newcommand{\Embi}[2]{\EMB{}_{#1}(#2)}
\def\Embprime#{\EMB{}'}
\def\Embpp#{\EMB{}''}
\newcommand{\Embprimew}[1]{\Embprime{}_{/\set{\v{#1}}}}

\def\DM#{\Delta}
\newcommand{\Dm}[1]{\DM{}(#1)}
\newcommand{\Dmi}[2]{\DM{}_{#1}(#2)}
\newcommand{\dm}[2]{y_{#1,#2}}

\def\DIFF#{\text{Diff}}
\newcommand{\Diff}[2]{\DIFF{}(#1,#2)}
\newcommand{\diff}[2]{\text{\emph{diff}}_{#1,#2}}
\newcommand{\diffs}[1]{\text{\emph{diff}}_{#1}}

\newcommand{\F}[1]{Feat(#1)}

\def\Fa#{Feat_{attack}}
\def\Ft#{Feat_{train}}
\newcommand{\f}[1]{f_{#1}}

\def\Adj#{Y}
\newcommand{\adj}[2]{\Adj{}_{#1,#2}}

\newcommand{\W}{W}
\newcommand{\w}[2]{w_{#1,#2}}

\newcommand{\Ng}[2]{\mathcal{N}_{#1}(\v{#2})}

\def\bins#{bins}
\newcommand{\bin}[1]{b_{#1}}

\newcommand{\set}[1]{\{#1\}}
\newcommand{\transpose}[1]{#1^T}
\newcommand{\abs}[1]{|#1|}
\renewcommand{\deg}[1]{deg(\v{#1})}
\newcommand{\degG}[2]{deg_{#1}(\v{#2})}

\newcommand{\argmax}[1]{\underset{#1}{\operatorname{arg}\,\operatorname{max}}\;}
\newcommand{\argmin}[1]{\underset{#1}{\operatorname{arg}\,\operatorname{min}}\;}

\def\fb#{Facebook}
\def\bb#{Barabasi}
\def\bbm#{Barabasi m5 n1000}

\def\ntwov#{node2vec}
\def\LINE#{LINE}

\newcommand{\mult}[2]{<#1,#2>}

\input{sections/00_abstract.tex}



\maketitle


\input{sections/01_introduction.tex}
\input{sections/02_related_work.tex}
\input{sections/03_approach.tex}

\input{sections/04_experiments.tex}
\input{sections/05_discussion.tex}
\input{sections/06_conclusions.tex}

\balance
\bibliographystyle{acm}
\bibliography{full_bib}

\input{sections/07_appendix.tex}

\end{document}

%% file: sections/00_abstract.tex
\begin{abstract}
        Data ownership and data protection are increasingly important topics with ethical and legal implications, e.g., with the \emph{right to erasure} established in the European General Data Protection Regulation (GDPR). 
    In this light, we investigate network embeddings, i.e., the representation of network nodes as low-dimensional vectors. 
    We consider a typical social network scenario with nodes representing users and edges relationships between them. 
    We assume that a network embedding of the nodes has been trained. After that, a user demands the removal of his data, requiring the full deletion of the corresponding network information, in particular the corresponding node and incident edges. 
    In that setting, we analyze whether after the removal of the node from the network and the deletion of the vector representation of the respective node in the embedding significant information about the link structure of the removed node is still encoded in the embedding vectors of the remaining nodes. If so, this would require a (potentially computationally expensive) retraining of the embedding. 
    For that purpose, we deploy an attack that leverages information from the remaining network and embedding to recover information about the neighbors of the removed node. 
    The attack is based on (i) measuring distance changes in network embeddings and (ii) a machine learning classifier that is trained on networks that are constructed by removing additional nodes. 
    Our experiments demonstrate that substantial information about the edges of a removed node/user can be retrieved across many different datasets. 
    This implies that to fully protect the privacy of users, node deletion requires complete retraining --or at least a significant modification-- of original network embeddings. Our results suggest that deleting the corresponding vector representation from network embeddings alone is not a sufficient measure from a privacy perspective.
\end{abstract}

%% file: sections/01_introduction.tex
\section{Introduction}\label{sec:introduction}
     \begin{figure}
        \centering
        \includegraphics[width=\linewidth]{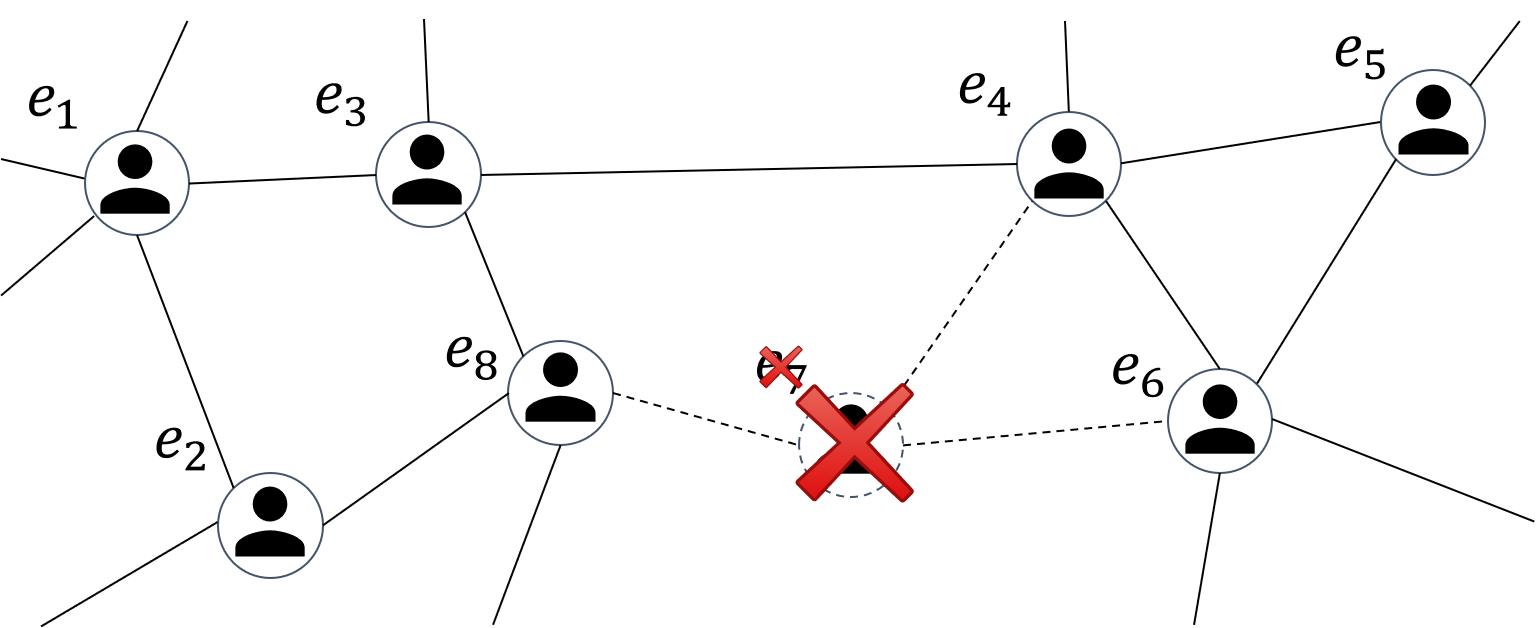}
        \caption{
        Scenario evaluated in this paper. Given a social network and a low dimensional node embedding on that network, a  user demands the deletion of his account. Can her neighborhood structure be restored from the embedding vectors of the remaining nodes?}
        \label{fig:task_illustration}
    \end{figure}

Network embeddings, i.e., the representation of network nodes as low-dimensional vectors, are a key technique of many state-of-the-art solutions for a broad range of social network analysis tasks such as link prediction~\cite{liben2007link,lu2011link}, node classification~\cite{bhagat2011node}, or community detection~\cite{fortunato2010community}. Since embedding calculation is often computationally expensive, recalculation after every change in a social network is not feasible for very large networks.
At the same time, data ownership and data protection become increasingly important topics for social media and social network applications. For example, the European General Data Protection Regulation (GDPR) establishes a \emph{right to erasure}, meaning that a data-holder is obligated to remove all personal data of an individual if requested. Previous work has demonstrated that machine learning models can leak information about the training data~\cite{sok,salem2018ml,veale2018algorithms}.  Similarly, network embeddings might open up the potential for residual personal data to remain in a system while explicit data regarding a user has been deleted.

\noindent\textbf{Research goal. }
We consequently investigate in this work the possibility of recovering private information from social network embeddings.
Specifically, we analyze the following scenario (illustrated in Figure~\ref{fig:task_illustration}):
We assume a social network with users as nodes and friendship relations as edges. On that network, an embedding with low-dimensional vector representations has been trained with a standard embedding algorithm (such as node2vec~\cite{grover2016node2vec} or HOPE~\cite{ou2016asymmetric}), e.g., to recommend new friends.
Now, one specific user deletes his account and requests the removal of his data, legally requiring the platform to delete all data on this user. 
In that setting, we study if the removal of the node from the network and deletion of the vector of the respective node is sufficient from a privacy perspective, or if significant neighborhood information of the removed node is still encoded in the embedding vectors of the remaining nodes, and a (potentially computationally expensive) retraining or modification of the embedding is required.
    
\noindent\textbf{Approach. }
   We develop a method\footnote{Code available at \url{https://github.com/embedding-attack/embAttack}} to extract link information about a removed user node from the embedding vectors of the other nodes: 
    Specifically, we train a new embedding $\EMB{}'$ on the remaining network without the removed node and calculate pair-wise differences in node similarities to the original embedding $\EMB{}$. These differences are used by a classifier to predict the neighbors of the removed user node. To train this classifier, we artificially remove additional nodes from the network, compute the respective embeddings $\EMB{}''$, and compare them with the embedding $\EMB{}'$.

    \noindent\textbf{Contribution and findings. }
    Our experiments demonstrate that substantial information about the edges of a removed node/user are leaked.
    For example, on a sample of a Facebook network, our method achieves an area under the ROC curve (AUC) score of $0.72$ and a precision@10 of $0.41$ when a \ntwov{} embedding is used, and an AUC of $0.91$ and a precision@10 of $0.6$ when a HOPE embedding~\cite{ou2016asymmetric} is utilized.
    This implies that to fully protect the privacy of users, node deletion requires complete retraining -- or at least a significant modification -- of the network embeddings.

%% file: sections/02_related_work.tex
\section{Related Work}
	Next, we discuss existing literature related to our work, in particular, network embeddings, link prediction, and privacy attacks on machine learning models.
	
	\noindent\textbf{Network embeddings. }\label{sec:related_work:Network Embedding}
		Network embeddings recently received strong attention from the research community due to their excellent performance for practical applications such as link prediction, node classification, visualization, clustering, or community detection.
		Thus, a large range of methods for the calculation of network embeddings has been proposed in literature including factorization-based methods ~\cite{balasubramanian2002isomap,yanardag2015deep,nikolentzos2017matching}, random walk-based methods~\cite{perozzi2014deepwalk,grover2016node2vec,liu2017semantic}, or Convolutional Neural Networks~\cite{kipf2016semi,niepert2016learning}.
		Although the performance of our approach might vary, it is, in principle, agnostic to the algorithms used for training the embedding. Thus, we do not elaborate further on the specifics of those approaches but refer to recent surveys on the topic~\cite{goyal2018graph,cai2018comprehensive}.

	\noindent\textbf{Link prediction. }
	    One specific application area of social network analysis is link prediction, i.e., the prediction of missing links between nodes in a network or these that are likely to occur in the future or were removed. See ~\cite{lu2011link,wang2015link} for an overview. 
	    Predicting links to a specific node can be used to identify private information in anonymized networks (e.g., \cite{cai2016collective,zheleva2007preserving}).
	    However, in link prediction typically information on the current network structure and/or on attributes of nodes (for which links are to be predicted) are exploited.  
	    Neither is available in our scenario for the removed node as we assume this information to be completely removed from the data. 
	    Thus, standard link prediction methods such as similarity-based methods~\cite{leicht2006vertex}, stochastic block models~\cite{white1976social}, or Probabilistic Entity-Relationship Models~\cite{heckerman2007probabilistic} cannot be applied.
		
	\noindent\textbf{Privacy attacks on machine learning models.}\label{sec:rel_work:priv_attack}
	    Due to increased overall awareness of privacy issues, the susceptibility of machine learning models 
	    to the extraction of personal information
	    has come into the scope of the research community (e.g., \cite{veale2018algorithms}).
		
		In that regard, one key attack type is \emph{model inversion attacks}.
		Here, given a traditional machine learning model (e.g., a classifier on tabular data), an adversary tries to infer values of sensitive attributes in the training data. Generally, attack scenarios can be divided into two variations \cite{sok}: \emph{White-box} attacks (e.g., \cite{ateniese2013hacking}) where the adversary has access to information about the model architecture and/or parameters, and \emph{black-box} attacks where the adversary does not have access to properties of the model.
		Black-box attacks are more common and relevant in terms of data privacy. In most scenarios, the adversary can use the model, i.e., apply inputs to the model and observe corresponding outputs (e.g., \emph{machine learning as a service} offers). 
		Many approaches evaluate different inputs to the model 
		on how likely they are part of the training data. If the input space is small enough, all possible inputs can be tested, for larger input spaces (domain-specific) gradient descent based approaches can be used \cite{modelInversion}.
		
		Attacks to detect if an input for a machine learning model was used during training are often referred to as \emph{membership inference attacks}~\cite{membershipInference,truex2018towards,salem2018ml}. The basic idea is to extract differences in the confidences of the outputs for data used during training and data that was not used.
		Advanced techniques, therefore, train \emph{shadow models}, which imitate the attacked model, on different generated training datasets.
		An attack classifier is trained on data from these shadow models and is applied to data of the attacked machine learning model to achieve the attack's goal.
		We see the privacy attack on network embeddings proposed in this work as the transfer of this general idea to the setting of network embeddings.

%% file: sections/03_approach.tex
\section{Approach}
    This section describes the proposed method. 
    First, we formally define the problem and then describe the main approach. 
    In the last part, we introduce a variation to the approach.
    
    \subsection{Problem Definition}\label{sec:problem_definition}
        In our scenario, we assume $\G{} = (V,E)$ to be an undirected, unweighted, and connected network of nodes $V$ (e.g., persons in a social network) and edges $E$ (e.g., friendship links between these persons) that is stored, controlled, and used by a specific data owner. Since persons often have unique friendship networks, 
        the links in a social network can potentially identify a person and should be considered personal information.
        We assume that (to improve offered services) the data owner computed a network embedding $\Emb{\G{}}$ of the network nodes, i.e., a mapping of each node to a fixed-dimensional vector of $d$ (e.g., $d=128$) real numbers\footnote{Note that we use the term \emph{network embedding} to denote a vector representation of each node of a network, not of the network itself as custom in conflicting terminology in literature}. 
        Our problem scenario (and attack approach) is in principle agnostic to the employed embedding algorithm, and we will use a variety of different state-of-the-art embedding algorithms such as \LINE{}~\cite{tang2015}, \ntwov{}~\cite{grover2016node2vec}, HOPE~\cite{ou2016asymmetric} and SDNE~\cite{wang2016structural} in our evaluation. 
        
        Further, we assume that on request of the respective user, the node $\v{i}$ (with all incident edges) is removed from the network and the vector corresponding to this node is removed from the embedding $\Emb{\G{}}$.
        We denote the remaining network as 
        $\Gprime{}=\Gw{i}=$ 
        $(\Vw{i},\E{}_{/\set{ \set{ \v{i},\v{j} }| \v{j}\in\V{}}} )$ and the embedding without the vector representation of that node as $\EMBw{i}=\Embw{\G{}}{i}$. Note that the embedding $\EMBw{i}$ has not been obtained by complete retraining, which might be computationally expensive or even unfeasible for each change in the network. Thus, it  does not contain explicit information on $v_i$ anymore, but implicit information may still be contained via embedding vectors of other nodes that have been influenced by previously existing links with node $v_i$.
        
        In this work, we investigate if an attacker (adversary) can recover information of the removed node $v_i$ by using the embedding of the remaining nodes and their network structure.
        The target is to predict incident edges (i.e., neighbors) of $\v{i}$ in $\G{}$.
        In particular, we will introduce and develop an attack algorithm $\mathcal{A}$ that predicts the neighbors of $\v{i}$ in $\G{}$. Formally, 
        $\mathcal{A}(\Gprime{},\EMBw{i}) \mapsto \Ng{\G{}}{i},$
        where $\Ng{\G{}}{i} = \set{\v{j}|\set{\v{i},\v{j}}\in \E{}}$ denotes the neighbors of $\v{i}$. 
        If able to do so, we show the general susceptibility of network embeddings to privacy attacks. 
        We assume that the embedding algorithm and parametrization of the algorithm are known to the attacker (\emph{``open code -- closed data''}), see also Section~\ref{sec:discussion}.
    
    \subsection{Privacy Attack}\label{sec:approach:main_approach}
        
        \begin{figure}
        	\centering
        	\includegraphics[width=\linewidth]{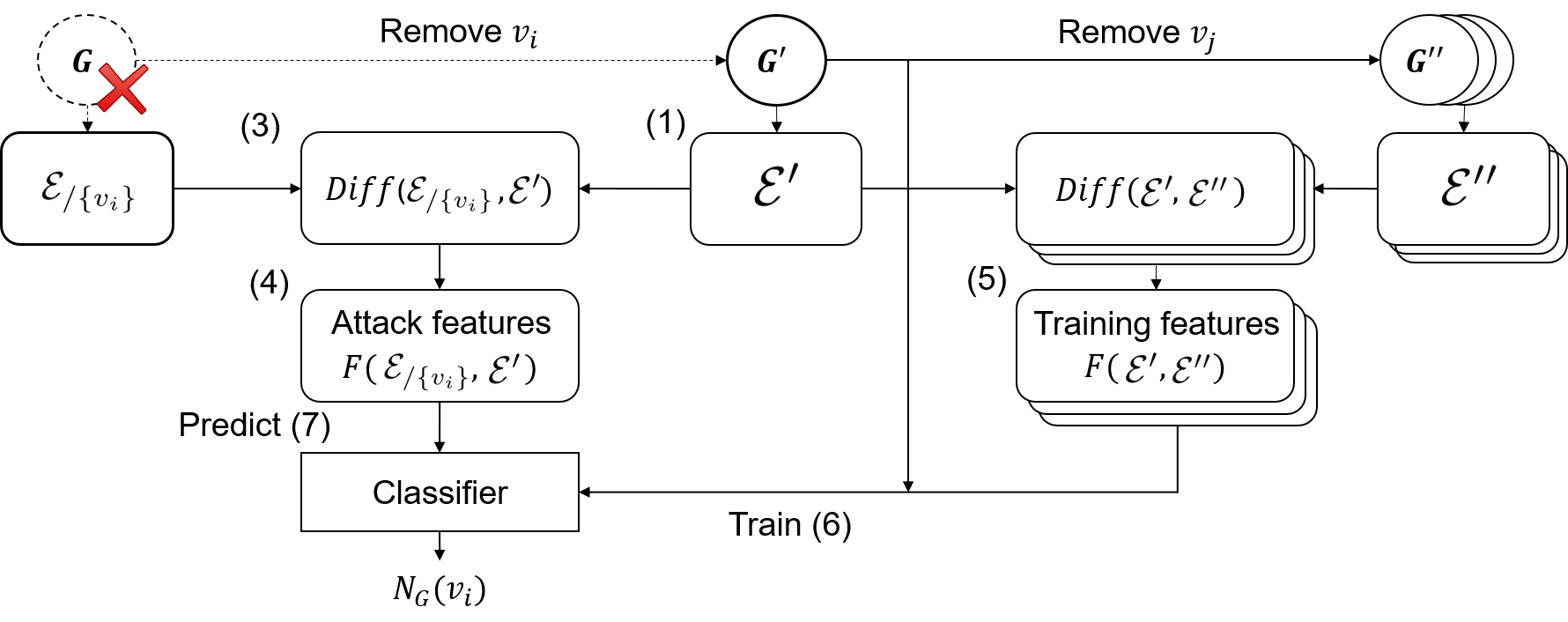}
        	\caption{Approach of the proposed attack. 
        	}
        	\label{fig:schema}
        \end{figure}
        Next, we describe our approach of recovering edge information of a deleted node $v_i$.
        The main idea is to develop an attack classifier that predicts for each node if it was connected to the removed node $\v{i}$ based on the differences in pairwise cosine distances of nodes between the original embedding $\EMBw{i}$ and a new embedding calculated on the network of the remaining nodes. To create training data for learning which distances are predictive for neighborhood with the removed node, we monitor the effects of removing additional nodes from the network.
        In detail, we proceed with the following steps, see Figure~\ref{fig:schema} for a visualization:
        \begin{enumerate}[leftmargin=*] 
            \item We start by computing a new embedding $\Embprime{}$ on the reduced network $\Gprime{}$ with the same algorithm and parametrization as the original one.
            \item From this embedding and the available embedding of the original network $\EMBw{i}$, we calculate respective distance matrices $\DM{}_{/\set{\v{i}}}$ and $\DM{}'$ that contain the pairwise distance between each node pair.
            \item Next, we  calculate the changes in distances between these two embeddings by calculating the (element-wise) difference matrix:
            	$\Diff{\EMBw{i}}{\Embprime{}} = \DM{}_{/\set{\v{i}}}-\DM{}'.$
        Thus, each element $\diff{k}{l}$ in $\Diff{\EMBw{i}}{\Embprime{}}$ describes the changes of the distance between nodes $\v{k}$ and $\v{l}$ from the original embedding $\EMBw{i}$ to the embedding of the remaining network $\Embprime{}$. Note that $\diff{k}{l} = \diff{l}{k}$ due to symmetry.
        
        Since most embeddings encode in some way a local neighborhood structure, intuitively nodes that exhibit many or large changes in the embedding distance are more likely to be close to the removed node. Initial experiments showed that this holds as a general trend, but a simple summation of taking maximum values is insufficient for a detailed prediction. Therefore, we use a machine learning approach to identify which distance changes are indicative for a node being a neighbor of the removed node.
        
        \item As a pre-requisite for a machine learning approach, we construct a feature vector $\f{k}$ for each node  $\v{k}\in\Gprime{}$ that characterizes the distribution of distance changes for this node in a histogram-like fashion. 
        For that purpose, we split the  range of all values in the full matrix $\Diff{\EMBw{i}}{\Embprime{}}$ (excluding the diagonal) by equal frequency discretization, i.e., we create bins $\bin{1},...,\bin{m}$ that contain the same number of values. 
        
        We now take the distance changes of $\v{k}$ to all other nodes $\set{\diff{k}{l}| \v{l}\in\Gprime{},\v{l}\neq\v{k}}$.
        For each bin, the number of these distance changes that lie within the bin then defines one node feature of feature vector $f_k$. We normalize these features over all bins.
        As an additional feature, we append the degree of $\v{k}$ in $\Gprime{}$, which is normalized over all features.
        
        \item Next, we create a training set to learn which features and feature combinations predict that a node was a neighbor of the removed node. For that purpose, we (temporarily) remove a second node $\v{j}$ from $\Gprime{}$ to obtain a network $\Gpp{}=\Gprimew{j}$ and compute the corresponding embedding $\Embpp{}$. Then we can compare the distance changes $\Diff{\Embprimew{j}}{\Embpp{}}$ between the embeddings  $\Embprimew{j}$ and $\EMB{}''$ and generate node features $f$ for these changes following steps (2)-(4). Additionally, for each node $\v{k}$, we can set a label $y_k$ that indicates if the respective node was a neighbor of the removed node $v_j$ in $\Gprime{}$.
        To obtain more training data, we can repeat this process for each node $v_j \in \Gprime{}$ as a node to be removed from $\Gprime{}$.
        
        \item Following that, we train a machine learning classifier with the data obtained in the last step, predicting the label $y_k$ (was a node a neighbor of the removed node) based on the features vector $f_k$ (with the distribution of distance changes and the degree) of the node $v_k$ in $\Gpp{}$. The intuition is that the changes for removing the second node in the network will lead to similar changes in the embedding distances compared to the changes induced by removing the first node $v_i$. 
        
        \item Finally, we use this classifier on the training features obtained from the network $\Gprime{}$ and embedding $\EMBw{i}$ to predict which nodes have been neighboring the removed node in the original network, i.e., which links the removed node had.
        
        \end{enumerate}

     \subsection{Variation of the Privacy Attack}\label{sec:approach:variations}
        In addition to the previously described approach, we also experiment with a variation. This variation aims to tackle the main challenge of the attack: Since most state-of-the-art embedding algorithms are based on approximate solutions to an optimization function (e.g., via gradient descent), the outcome of an embedding algorithm is not deterministic but varies to some degree even if the same algorithm is used on the exact same data. Thus, a part of the differences that we encounter when comparing embeddings of two different networks is only due to instabilities in the embeddings and not due to changes in the network. 
        
        The variation attempts to reduce the 
        effect of those instabilities. 
        To compute attack features, we calculate a number of embeddings ($\EMB{}^{'}_1,...,\EMB{}^{'}_m$) on the \emph{available} network $\Gprime{}$. 
        From those embeddings, we choose only the one most similar to the available embedding of the attacked network $\EMBw{i}$ for the computation of attack features.
        As distance measure between the embeddings we use the absolute sum over all values in the difference matrix $d(\EMBw{i},\EMB{}^{'}_k) = sum(abs(\Diff{\EMBw{i}}{\EMB{}^{'}_k}))$.
        The idea of this distance measure is that most node distances in the original embedding should be very similar to the node distances in the reduced embedding
        as the underlying networks only differ by one node.
        Training data is generated analogously, where the chosen embedding for the computation of attack features $\Embprime{}$ is used for training feature computations.

%% file: sections/04_experiments.tex
\section{Experimental Setup}\label{sec:experimental_setup}
This section describes the general experimental approach, including parameter settings and evaluation criteria, as well as datasets and embedding algorithms used in the evaluation.

    \subsection{Experimental Approach}
	To simulate an attack scenario, we compute an embedding on the original network $\G{}$
	and then remove a node $\v{i}$ from the network and its corresponding vector from the embedding.
	The network without the removed node $\Gw{i}=\Gprime{}$ and the embedding without the removed node $\Embw{\G{}}{i}=\EMBw{i}$ is then given as information for the attack. Results are evaluated using neighbors of $\v{i}$ in $\G{}$.
	On all datasets, we perform $15$ attacks, each with a different removed node $\v{i}$. Since the degree of a node has a significant effect on the performance of the attack (see Section~\ref{sec:exp:node_degree}), we sample $5$ nodes with low degrees (i. e. with degrees close to the minimum degree, usually $1$), $5$ with medium degrees, and $5$ nodes with high degrees. 
	Each experiment is repeated $5$ times on the same network and removed node, but with different embedding runs.
	Since the standard error over these repetitions is mostly $< 0.01$ and consistently $< 0.05$ for the reported measures, we only report mean values of the outcomes for conciseness.
	Experiments on the variation use the same embeddings on the available networks $\Gprime{}$ and $\Gpp{}$ for each iteration to decrease computational efforts.

	\noindent\textbf{Parameter Settings. }
    If not otherwise mentioned, we use $10$ different training networks $\Gpp{}$ for each attack and use $10$ bins for the creation of features, see also Section~\ref{sec:parameters} for an evaluation of parameter sensitivity. We further compare several different classifiers:
    \emph{K-nearest neighbor}, \emph{support vector machines}, \emph{decision trees}, \emph{random forests}, \emph{AdaBoost} using decision trees, and \emph{Gaussian naive Bayes}.
	We use standard parametrizations of the \emph{scikit-learn} python package~\cite{scikit-learn}.
	In most experiments we utilize the Gaussian naive Bayes classifier due to its solid performance, see also Section~\ref{sec:exp:classifier}.
	
    \noindent\textbf{Evaluation measures. }
    In terms of evaluation measures, we treat our setting as a (link) prediction problem and use established measures for this task. In particular, we employ the \emph{area under the ROC curve} (\emph{AUC}) of the prediction of neighbors to the removed node, the \emph{precision@10}, (i.e., the precision among the 10 candidates with the highest predicted likelihood of being a neighbor of the removed node), as well as the $F_1$-score of the prediction. For the $F_1$-score, we distinguish between averaging 
    over all attack scenarios (\emph{Macro-$F_1$}) and over (to predict) incident edges of the removed nodes in all scenarios (\emph{Micro-$F_1$}).

    \subsection{Data}
    We evaluate the proposed attack on several artificial and real-world networks. Due to resource limitations, we only conducted experiments on networks of $1000$ to $2000$ nodes, see also Section~\ref{sec:discussion}. 
    For larger networks, we use snowball sampling to extract subnetworks with $2000$ nodes\footnote{Samples are available at \url{https://github.com/embedding-attack/embAttack}}.
    In terms of datasets, we test our approach on a mix of synthetic and real-world networks:

    \noindent\textbf{\bb{}.} 
    To generate synthetic datasets, we use the well-known Barabasi-Albert model~\cite{barabasialbert1999} for scale-free networks. 
    The model has parameters to modify the number of nodes in the network and the number of edges relative to the number of nodes (\emph{parameter of attachment}) that control network size and density. 
    Unless otherwise mentioned, we use a generated network with $1000$ nodes and a parameter of attachment of $5$, which we refer to  as \bb{}.
    
    \noindent\textbf{\fb{}.} 
    This network is a subset of the friendship network of Facebook\footnote{\url{http://konect.uni-koblenz.de/networks/facebook-wosn-links}}. A node represents a user and an edge between two nodes represents a friendship between the users.
    We use snowball sampling to sample a network with $2000$ nodes and $14251$ edges.
    
    \noindent\textbf{Hamsterster.}
    Hamsterster was a social network for people who like hamsters\footnote{\url{http://konect.uni-koblenz.de/networks/petster-friendships-hamster}}. It operated for $10$ years but is shut down now. 
    We use the biggest connected component of this network, which has $1788$ nodes and $12476$ edges.
    
    \noindent\textbf{DBLP.}
    This network is a collaboration network extracted from the DBLP computer science bibliography~\cite{DBLP:journals/corr/abs-1205-6233}. Nodes correspond to authors. Two nodes are connected if the two authors collaborated in at least one paper. The used network is snowball sampled. It has $2000$ nodes and $7036$ edges.
    
    \subsection{Embedding Algorithms}
    We evaluate our attack on four embedding algorithms that we consider as representatives for some fundamental variations of computing network embeddings:
    \emph{HOPE}~\cite{ou2016asymmetric} embeds the neighborhood structure for each node using the Katz index.
        \emph{LINE}~\cite{tang2015} embeds the local and global neighborhood structures separately and combines the resulting embeddings. \emph{node2vec}~\cite{grover2016node2vec} applies the word embedding algorithm \emph{word2vec} on random walks generated from the network. 
        The Structural Deep Network Embedding (\emph{SDNE})~\cite{wang2016structural} computes embeddings based on the encoder-decoder principle. 
    %
    Among those algorithms, two focus on global neighborhood structure (HOPE and SDNE with a low alpha parameter), while the two others (\LINE{} and \ntwov{}) balance embedding local and global neighborhoods.
    Additionally, HOPE appears to yield near-constant embeddings, while the other algorithms display significant instability (that is, variations in the embeddings between algorithm runs on the same data).
    
    To enable an unbiased comparison between the embeddings from these algorithms, we use a dimension of $128$ for each.
    For other parameters, standard values that are either proposed by the original papers or are approved parameter settings are used.
    While this may lead to sub-optimal performances, we expect the attack performance to only increase with optimized parameterization as more information will be encoded in the embedding.

\section{Experimental Results}
	This section presents the results of our experiments. First, we present key overall results on the performance of our attack approach on several different networks and embedding algorithms. 
    Then, we analyze the effect of network size, network density, and node degree on the attack. Afterwards, we evaluate the sensitivity to parameter choices in our approach. 
    Finally, we investigate 
    the influence of instability in the embedding computations on the attack and
    the variation of the main attack, described in Section~\ref{sec:approach:variations}, which attempts to reduce the influence. 
    
        \begin{table}
            \caption{Performance of the attack on different networks and algorithms. The attack extracts some information from all embeddings on most networks. There is no embedding algorithm the attack clearly performs best on.}
             \small\centering \begin{tabular}{llcccc}
                \toprule
                Networks     &          &        HOPE &        \LINE{} &    \ntwov{} &        SDNE \\
                        \midrule Barabasi & AUC &  0.63 &  0.83 &     0.72 &  0.61 \\
                               & precision@10 &  0.34 &  0.51 &     0.36 &  0.33 \\
                               & Macro-$F_1$ &  0.16 &   0.4 &     0.19 &  0.16 \\
                               & Micro-$F_1$ &  0.17 &  0.47 &     0.19 &  0.14 \\
                        \midrule Facebook & AUC &  0.91 &   0.7 &     0.72 &  0.73 \\
                               & precision@10 &   0.6 &  0.45 &     0.41 &  0.42 \\
                               & Macro-$F_1$ &  0.34 &  0.33 &     0.17 &  0.22 \\
                               & Micro-$F_1$ &  0.29 &   0.2 &      0.2 &  0.15 \\
                        \midrule Hamsterster & AUC &  0.83 &  0.69 &     0.66 &  0.72 \\
                               & precision@10 &  0.39 &  0.29 &     0.27 &  0.25 \\
                               & Macro-$F_1$ &  0.19 &  0.23 &     0.19 &  0.18 \\
                               & Micro-$F_1$ &  0.14 &  0.24 &     0.22 &  0.18 \\
                        \midrule DBLP & AUC &  0.96 &  0.74 &      0.6 &  0.65 \\
                               & precision@10 &  0.16 &  0.14 &     0.08 &  0.07 \\
                               & Macro-$F_1$ &  0.19 &  0.11 &     0.06 &  0.08 \\
                               & Micro-$F_1$ &  0.21 &  0.12 &     0.05 &  0.06 \\
                \bottomrule
            \end{tabular}

            \label{tab:main_results}
        \end{table}
        
    \subsection{Main Results}\label{sec:exp:main_results}
        To tackle our main research question, i.e., whether information about the neighborhood of a removed node can be recovered using the vector representations of the other nodes, we evaluate the attack on several different networks and embedding algorithms. Our experimental results, see Table~\ref{tab:main_results}, show that the attack can recover substantial information on the neighborhood of the removed nodes with decent success on many networks from all evaluated algorithms. While the recovered information is often not enough to predict the neighbors accurately, the approach definitely recovers good, partially excellent, candidates for neighbors of the removed node as indicated by high AUC values of up to $0.96$. This might, in practical situations, already be enough to identify an individual.
    	
    	There are also strong variations in the performance of the attacks. In general, there is no embedding algorithm for which the attack performs best over all networks.
    	A possible explanation is that different algorithms vary in behavior on different network structures. On the regular structured \bb{} network, the attack performs best on \LINE{}, whereas on less regular structured networks such as \fb{} the attack performs best on HOPE.
    	The better performance on these embeddings may be due to less instability in their computations, which makes the comparison between embeddings more consistent, see Section~\ref{sec:exp:variation}.
        The performance of the attack varies with different networks, and even different measures lead to varying conclusions. According to AUC, neighbors can be predicted best in \fb{} and DBLP. A substantial amount of correct predictions (according to precision@10 and the $F_1$-scores) can be achieved for the \bb{}, \fb{}, and Hamsterster networks. 
       Overall, it is possible to extract substantial information, but only seldom precise neighborhoods with all network embeddings, but this is subject to strong variations. 
       Hence, we next investigate the effect of various conditions on the performance of the attack.

   \subsection{Impact of Structural Properties} 
           	\begin{figure} [t!]
    		\begin{subfigure}{0.99 \linewidth}%
    	        \includegraphics[width=\linewidth]{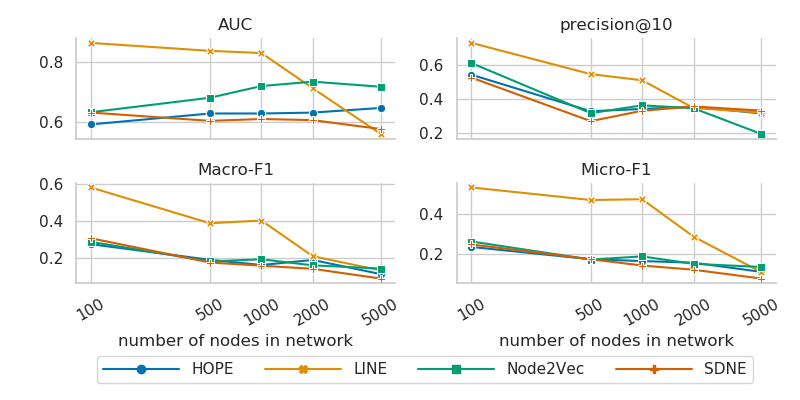}
        	    \subcaption{Attack performance with varying network size}
        	    \label{fig:exp:graph_size}
        		\centering
        	\end{subfigure}
        	
    	   \begin{subfigure}{0.99 \linewidth}%
    	        \includegraphics[width=\linewidth]{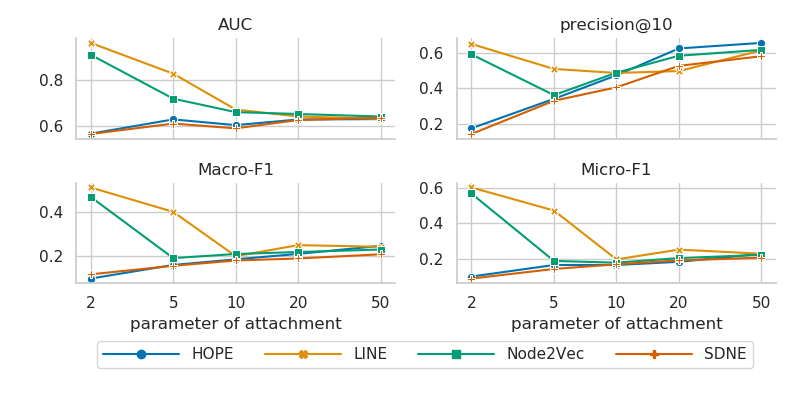}
        	    \subcaption{Attack performance with varying density}
        	    \label{fig:exp:graph_density}

        		\centering
        	\end{subfigure}%
        	\caption{Performance of the attack on networks generated by the Barabasi-Albert model with different network sizes and parameters of attachment (densities). Attack performance declines with network size in a majority of cases. 
        	On low density networks \LINE{} and \ntwov{} are more susceptible to attacks.}
    	\end{figure}

        Here, we evaluate the effect of network size and density on the performance of the attack. We further investigate differences based on the degree of the attacked node.
	    
	    \subsubsection{Network size.}
	    To evaluate the impact of network size on attack performance, we generated Barabasi-Albert networks with varying sizes and a fixed parameter of attachment of $5$.
    	
    	Our results (see Figure~\ref{fig:exp:graph_size}) show that according to AUC, increasing network size has a strong 
    	negative effect on the performance of attacks on the \LINE{} embedding, but little influence for attacks on \ntwov{}, HOPE, and SNDE.
        According to other evaluation criteria, performance declines for all embeddings on larger networks. 
        Keeping vector representations at a fixed size might lead to decreased performance since an increased amount of information must be captured in the same number of dimensions.
        Overall, making precise predictions of the removed node's neighbors gets more challenging with increased network size. However, a set of candidates for such neighbors can still be extracted from many embeddings.
        
    	\subsubsection{Network density.}
    	Next, we analyze the effect of network density on attack performance. For that purpose, we generate multiple networks with the Barabasi-Albert model with a fixed size of $1000$ nodes, but varying attachment parameter. 
    	
    	The results (see Figure~\ref{fig:exp:graph_density}) indicate that \LINE{} and \ntwov{} are substantially more susceptible to attacks for low-density networks compared to the other algorithms according to all measures. This might be due to the higher emphasis on the local neighborhood of these embeddings.
    	There is also a clear trend that the precision@10 increases for parameters of attachment $>5$.
    	This may be due to quantitatively stronger differences between $\EMB{}$ and $\Embprime{}$ that result from the removal of the node, but also just from the fact that there are more positives in the data set.

	\subsubsection{Node degree}\label{sec:exp:node_degree}
	    \begin{table}
            \caption{Performance of the proposed attack on sets of nodes with low, medium, and high degrees. The performance of the attack is significantly better for nodes with higher degrees.}
            \label{tab:exp:degree}
            \setlength\tabcolsep{1pt}
            \small\centering \begin{tabular}{llcccccccccccc}
            \toprule
			        &          & \multicolumn{3}{c}{Barabasi} & \multicolumn{3}{c}{Facebook} & \multicolumn{3}{c}{Hamsterster} & \multicolumn{3}{c}{DBLP} \\
			       \cmidrule(lr){3-5} \cmidrule(lr){6-8}\cmidrule(lr){9-11} \cmidrule(lr){12-14}
				       &          &      low & med. &  high &      low & med. &  high &         low & med. &  high &   low & med. &  high \\
                \midrule \parbox[t]{2mm}{\multirow{3}{*}{\rotatebox[origin=c]{90}{HOPE\text{   }}}} & AUC &     0.65 &    0.6 &  0.64 &        1 &   0.86 &  0.86 &           1 &   0.72 &  0.78 &     1 &   0.94 &  0.95 \\
                       & pr@10 &     0.04 &    0.4 &  0.58 &     0.12 &   0.84 &  0.84 &         0.1 &   0.57 &   0.5 &  0.06 &    0.1 &  0.32 \\
                       & Macro-$F_1$ &     0.04 &   0.14 &  0.21 &     0.03 &   0.35 &  0.35 &        0.02 &   0.15 &  0.19 &  0.05 &   0.21 &  0.29 \\
                       & Micro-$F_1$ &     0.04 &   0.14 &  0.21 &     0.02 &   0.36 &  0.34 &        0.02 &   0.15 &  0.18 &  0.02 &   0.21 &  0.32 \\
                \midrule \parbox[t]{2mm}{\multirow{3}{*}{\rotatebox[origin=c]{90}{LINE\text{  }}}} & AUC &     0.71 &   0.89 &  0.89 &      0.7 &   0.67 &  0.72 &        0.61 &   0.75 &  0.72 &  0.68 &   0.76 &  0.79 \\
                       & pr@10 &     0.04 &   0.59 &  0.89 &     0.08 &   0.63 &  0.65 &        0.04 &    0.5 &  0.34 &     0 &   0.32 &  0.09 \\
                       & Macro-$F_1$ &     0.07 &   0.47 &  0.57 &     0.05 &   0.21 &  0.23 &        0.01 &   0.31 &  0.27 &  0.01 &   0.19 &   0.1 \\
                       & Micro-$F_1$ &     0.07 &   0.48 &  0.57 &     0.04 &   0.21 &  0.23 &        0.01 &    0.3 &  0.27 &  0.01 &   0.19 &  0.11 \\
                \midrule 
                \parbox[t]{2mm}{\multirow{3}{*}{\rotatebox[origin=c]{90}{node2vec\text{  }  }}} & AUC &      0.7 &   0.75 &  0.71 &     0.72 &    0.7 &  0.72 &        0.51 &   0.75 &  0.73 &   0.6 &   0.59 &  0.61 \\
                       & pr@10 &     0.08 &    0.4 &  0.61 &        0 &   0.62 &   0.6 &           0 &   0.49 &  0.33 &  0.02 &   0.14 &  0.07 \\
                       & Macro-$F_1$ &     0.07 &   0.21 &  0.21 &        0 &   0.21 &  0.25 &           0 &   0.29 &  0.29 &  0.01 &    0.1 &  0.06 \\
                       & Micro-$F_1$ &     0.06 &   0.21 &  0.21 &        0 &   0.21 &  0.25 &           0 &   0.28 &  0.28 &  0.01 &    0.1 &  0.05 \\
                \midrule \parbox[t]{2mm}{\multirow{3}{*}{\rotatebox[origin=c]{90}{SDNE\text{    }}}} & AUC &     0.64 &   0.59 &   0.6 &     0.77 &   0.72 &  0.69 &        0.66 &   0.76 &  0.72 &   0.6 &   0.71 &  0.65 \\
                       & pr@10 &     0.06 &   0.37 &  0.56 &     0.03 &   0.63 &  0.59 &           0 &   0.42 &  0.32 &     0 &   0.13 &  0.06 \\
                       & Macro-$F_1$ &     0.06 &   0.15 &  0.17 &     0.02 &   0.19 &  0.19 &           0 &   0.23 &  0.21 &  0.02 &   0.09 &  0.08 \\
                       & Micro-$F_1$ &     0.06 &   0.14 &  0.17 &     0.01 &   0.18 &  0.18 &           0 &   0.23 &  0.21 &  0.02 &   0.07 &  0.08 \\
                       
            \bottomrule
            \end{tabular}
        \end{table}
	
        \begin{figure*} [t!]
		 \begin{subfigure}{0.99 \linewidth}%
	        \includegraphics[width=\textwidth]{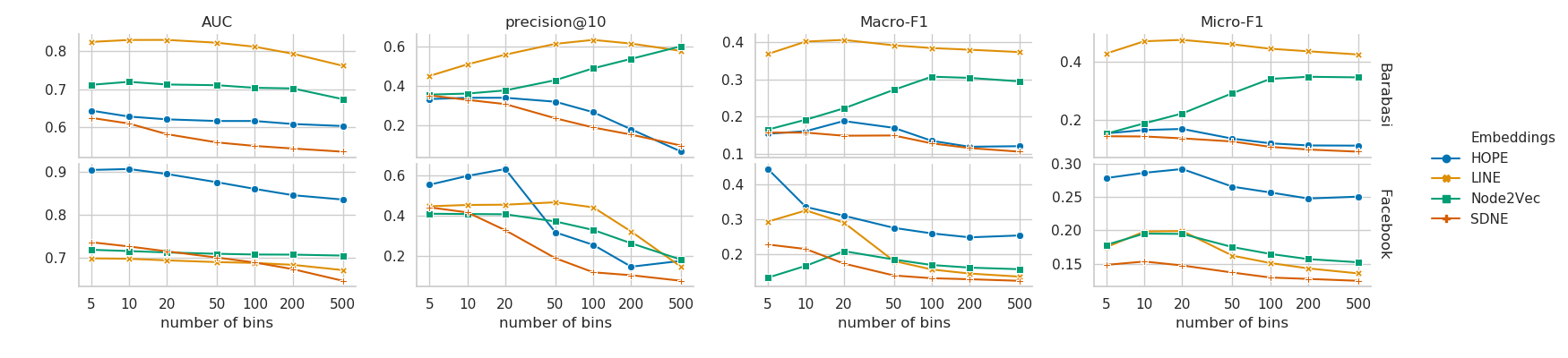}
    	    \subcaption{Attack performance for different number of bins for feature construction}
    	    \label{fig:exp:num_bins}
    		\centering
    	\end{subfigure}%
		
		\begin{subfigure}{0.99 \linewidth}%
	        \includegraphics[width=\linewidth]{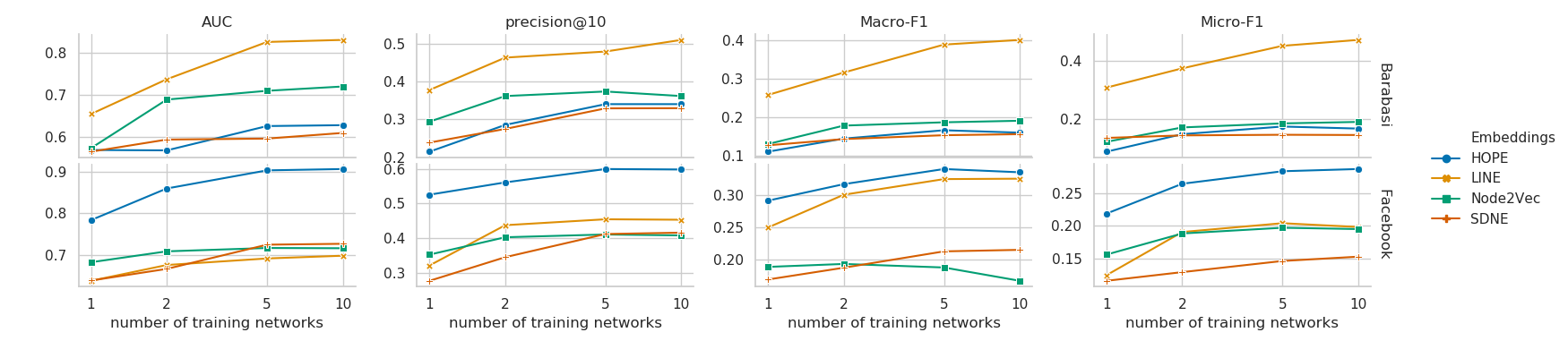}
    	    \subcaption{Attack performance with different numbers of training networks ($\Gpp{}$)}
    	    \label{fig:exp:num_tr_graphs}
    		\centering
    	\end{subfigure}
    
    	\caption{Parameter sensitivity of attack performance. We investigate attacks with 
    	different numbers of bins, for feature construction, and training networks ($\Gpp{}$), to generate training data, on the \bb{} network (top rows) and the \fb{} network (bottom rows). The optimum number of bins varies with embeddings, networks, and evaluation criteria. More training data improves the performance of the attack, but mostly plateaus between 5 and 10 networks.}
	\end{figure*}
	
	    As described in Section~\ref{sec:experimental_setup}, we sample attacked nodes with low, medium, and high degrees. Differences between the types of nodes are shown in Table~\ref{tab:exp:degree}.
		For the attacks on nodes with a low degree, the data clearly shows 
		that precision@10, Micro-$F_1$ and Macro-$F_1$ are very low for all embedding algorithms and networks. Thus, predicting neighbors of low degree nodes exactly is overall very challenging. 
		Note that AUC values vary between very low and high values, probably due to the small number of positives in these cases.
		On the other hand, neighborhoods of medium and high degree nodes can be predicted substantially better, reaching precision@10 of over $0.8$ on the \fb{} network and $0.5$ on Hamsterster.
		An explanation for this is that low degree nodes usually have a smaller impact on the vector representations of their neighbors.

    \subsection{Parameter Settings}
	\label{sec:parameters}
    	    This section presents experiments on the sensitivity to parameter settings of the attack, i.e., the number of bins for training feature construction, the number of training networks, and the selection of the classifier. For space reasons, we limit the presented results to two networks, namely the Barabasi and Facebook networks.
    	    
    		\subsubsection{Number of bins.}
          		First, we evaluate the influence of the number of bins used for the computation of the training features.
        		Figure~\ref{fig:exp:num_bins} shows the performance of the attack over a range of different numbers of bins. 
        		Overall, we see that the optimal number of bins varies significantly depending on the dataset, the embedding algorithm, and the measure used for evaluation.
        		On both networks, the AUC values are highest for  low numbers of bins and decline slightly with an increasing number of bins, probably due to over-fitting. 
        		In contrast, according to precision and $F_1$-scores, high values are to be preferred specifically for the \bb{} network and the \ntwov{} and \LINE{} embeddings, while low numbers of bins (of up to 20) are better on the Facebook network and the HOPE embedding.
        		We conclude that, if desired, the number of bins can be fine-tuned to a specific scenario depending on the concrete network and embedding method.

    		\subsubsection{Number of training networks.}
            	Next, we evaluate the effect of the number of training networks $\Gpp{}$ on the attack performance. In general, higher numbers of networks provide more training data but substantially increase the runtime since a separate embedding must be trained for each network $\Gpp{}$.
            	Results, see Figure~\ref{fig:exp:num_tr_graphs}, show that using multiple training networks leads to better classifications, but peak performance is reached at $5$ training networks in most cases. We have set the number of training networks used in the other experiments accordingly. We expect a very slight increase in performance for more training data, but at the cost of substantially more computational efforts, see also Section~\ref{sec:discussion}.
            	
            \subsubsection{Classification algorithms.}\label{sec:exp:classifier}
                Finally, we evaluate several classifiers for the attack. In particular, we evaluate  \emph{K-nearest neighbor classification}, \emph{support vector machines}, \emph{ decision trees}, \emph{random forests}, \emph{AdaBoost}, and \emph{Gaussian naive Bayes classifiers}. 
                The outcomes vary substantially, see the Appendix for detailed results. 
                The overall best performing and most consistent classifier is the \emph{Gaussian naive Bayes classifier}.
                \emph{Support vector machines} generally perform poorly.
                \emph{K-nearest neighbor classification}, \emph{decision trees}, and \emph{random forests} only learn decently on the \LINE{} embedding of the synthetic \bb{} network and on HOPE embeddings of real-world networks. In these cases, \emph{AdaBoost}, which uses decision trees, performs comparably to (and in some cases outperforms) the Bayesian classifier on all evaluation criteria. For example, when embedding network \fb{} with HOPE, AdaBoost reaches a precision@10 of $0.73$ and the Bayesian classifier of $0.6$. Using other embedding methods and networks, AdaBoost achieves comparable, often better, performances according to AUC,
                but worse according to other evaluation criteria. E.g., when embedding network Hamsterster with \LINE{}, AdaBoost reaches an AUC of $0.75$ but a precision@10 of only $0.1$ and a Micro-$F_1$ score of almost $0$ while the Bayesian classifier reaches an AUC of $0.6$, precision@10 of $0.29$ and a Micro-$F_1$ score of $0.24$.
                Thus, we use the Gaussian naive Bayes classifier as a consistent choice in the other experiments for this paper, but again see options for fine-tuning a particular attack.

    \subsection{Instability in Embedding Algorithms}\label{sec:exp:variation}
  
        As explained in Section~\ref{sec:approach:variations}, random effects in embedding computations influence the performance of the attack. We evaluate these influences and the described variation to the attack to reduce them.
        
\begin{figure*} [t!]
		 \begin{subfigure}{0.99 \linewidth}%
	        \includegraphics[width=\textwidth]{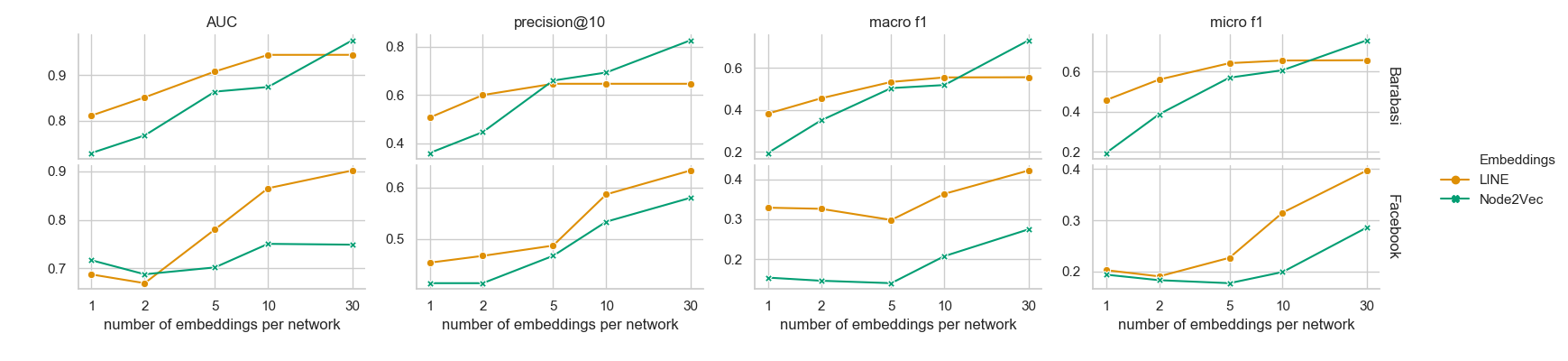}
    	   \vspace{-2mm}
    	    \subcaption{Average over distance matrices from multiple embeddings per network}
    	    \label{fig:exp:average_over_embeddings}
    		\centering
    	\end{subfigure}%
		
		\begin{subfigure}{0.99 \linewidth}%
	        \includegraphics[width=\linewidth]{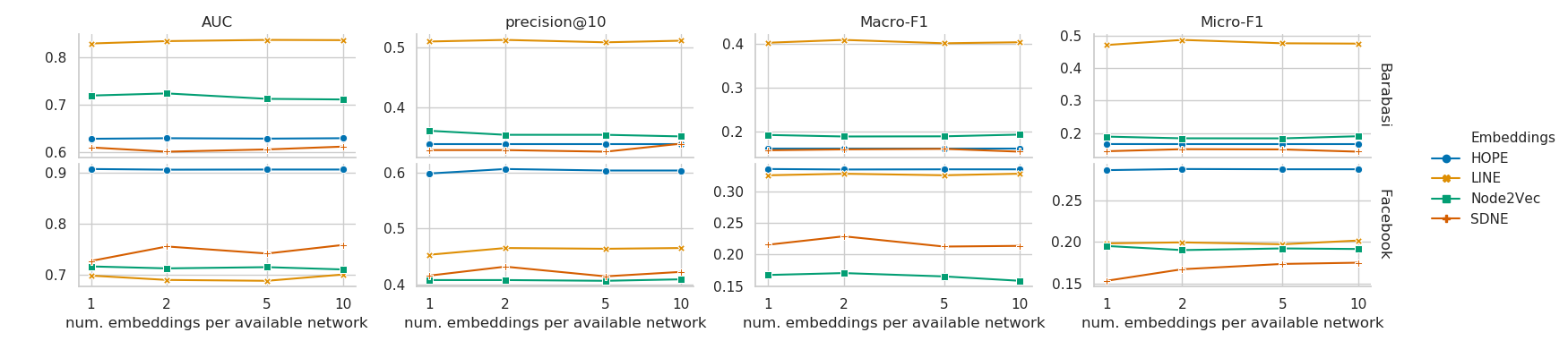}
    	    \subcaption{Variation of the attack using different numbers of embeddings on available networks ($\Gprime{}$ and $\Gpp{}$).}
    	    \label{fig:exp:most_sim_embeddings_multiple_embeddings}
    		\centering
    	\end{subfigure}
    	    \caption{Influence of random factors in the embedding calculations and performance of the variation on networks \bb{} (top rows) and \fb{} (bottom rows). 
    	    We reduce the influence of random effects in embedding computations by averaging pairwise node distances 
    	    over multiple embeddings per network. Fewer random effects increase the attack performance significantly (a). A variation to the approach, which computes multiple embeddings on available networks ($\Gprime{}$ and $\Gpp{}$) and selects the most similar to the attacked embedding, does not yield significant improvements (b).
    	    }
	    \end{figure*}

       To analyze the influence of instabilities, we reduce this influence by averaging for each network
       over distance matrices computed from multiple embeddings on that network. The resulting average distance matrices are used like normal distance matrices in the main approach.
        Figure~\ref{fig:exp:average_over_embeddings} shows the performance of the attack using different numbers of embeddings per network on embedding algorithms \LINE{} and \ntwov{} and networks \bb{} and \fb{}. The performance of the attack increases with decreasing influence of random effects (i.e., more embeddings per network). For \bb{} the performance increase on embedding \LINE{} stagnates at about $10$ embeddings per network, but further increases on \ntwov{}.
        The attacks on \fb{} also benefit greatly.
        This supports the hypothesis that the 
	 	instabilities in the computation of the embeddings are a major limiting factor for the performance of the attack. Overcoming this would significantly improve the performance.
	 	
	 	 \noindent\textbf{Variation.}
    	 Section~\ref{sec:approach:variations} describes an attempt to reduce the influence of these effects on our attack approach. Experimental evaluation of this approach (see Figure~\ref{fig:exp:most_sim_embeddings_multiple_embeddings}) shows that this approach in its basic form does not improve attack performance. A likely problem is that the employed measure for embedding similarity is insufficient.
        We can even observe that in some cases worse embeddings are chosen, as the performance decreases with more embeddings per network. Future research in that direction is warranted.

%% file: sections/05_discussion.tex
\section{Discussion}\label{sec:discussion}

    In this section, we present the implications of the findings and evaluate limitations of the approach.
    

    %
    
\subsection{Implications}
    Our results show that after removing a node from a network and its respective vector representation from the network embedding, substantial information on the neighborhood of this node remains in the embedding. Even though the exact neighbors cannot be recovered reliably, several ``likely friends'' can often be found. This might be enough in many scenarios to identify the person behind the removed node. Further, this can lead to the revelation of critical private information, such as political or religious affiliation~\cite{he2006inferring, lindamood2009inferring}.
    
    Potentially, this can  have crucial implications: Assume, for example, that a repressive governance entity foreign to the social network maintainer successfully obtains a snapshot of the network and an embedding illegally. It could then identify an individual by recovering some of his friends. At the same time, another recovered edge in the network could show contact to another network node, which represents a political dissident. In this case, the failure of erasing the full information of a node to be deleted in the network embedding could lead to strong real-life ramifications such as political persecution.
    
    
    This has consequences for the usage of network embeddings
    in the context of a typical social network scenario. 
    Whenever a user demands the deletion of his network information, it is not sufficient to remove the corresponding node from the network and the corresponding vector representation of that node from embeddings trained on the network to fully protect her privacy. To ensure that all his data is removed, retraining or at least modification of the remaining embeddings is required. This can be computationally expensive and has to be repeated whenever any user deletes his account. Even if this measure might be technically simple to accomplish, it might be computationally costly. 
    Moreover, the platform must explicitly be aware of this potential privacy issue. 
    In that direction, our paper raises awareness of possible privacy concerns in the context of non-updated network embeddings.
    
    With the increasing importance of data ownership and data protection, our results might also  have legal implications:
    As the GDPR establishes a right to erasure, on a person's request a company is legally required to remove all personal data. This includes data that can lead to the identification of that person and should -- by the results of this work -- also involve network embeddings.

\subsection{Limitations}
    We see our work as a first step in the analysis of privacy attacks on embeddings that is still restricted in multiple ways:

    As described above, due to resource limitations we only evaluate the attack on relatively small networks.
    The attack can, however, be parallelized and thus scales well with increased computational resources, enabling application also to larger networks. Also, a privacy attack can be a one-time effort, while the platform maintainer would be required to recompute network embeddings on a regular basis.
    Additionally, small and specialized social networks are also heavily used in many domains, and those small networks might be specifically vulnerable to security attacks.
    
    We only evaluated the performance on connected networks, while many real-world (especially social) networks consist of multiple components. 
    The attack may work for such networks by analyzing each component separately as long as the removed node did not connect components.
    Also, to establish that information is preserved, it was sufficient to only look at one removed node at a time. 
    In real-world scenarios, the deletion of multiple nodes may be common, making a successful attack more challenging. 
    Our approach can be extended to that scenario in the future, e.g., by
    applying the proposed method repeatedly.
    Additionally, we assume that the attack algorithm knows the applied embedding algorithm and its parameterization. While this is a strong assumption, it appears to be plausible either through security leaks, published research, or an open source implementation of the online social network. Limiting this assumption will be another interesting direction for future research.
    
    Our approach recovers network information with high, but far from perfect accuracy. Yet, combining its results with other sources might provide legally critical information on individuals. While we expect that our experimental results could be further improved by optimizations, we do not expect that a full recovery of the network structure is plausible given that each embedding always represents only a lossy compression of the original network.
    %
    In our experiments, we see strong variances in the susceptibility of embeddings to privacy attacks. An interesting open question is if this correlates with the performance of the embeddings on other tasks. We see these open issues as promising directions for future research.


%% file: sections/06_conclusions.tex
\vspace{-1mm}
\section{Conclusions and Future Work}
    We studied the susceptibility of social network embeddings to privacy attacks. We asked if it is sufficient to remove a node from a network and the corresponding vector representation from the embedding trained on this network to delete personal information on that node, specifically if the neighborhood of the removed node can be recovered with information encoded in the embedding vectors of the remaining nodes. 
    For that purpose, we deployed an attack based on differences between node distances in network embeddings and a machine learning classifier.
    Our experimental results demonstrate that substantial information about the edges of a removed node/user can be retrieved across many different datasets. We also present insights into which kind of nodes are particularly susceptible to these kinds of attacks and find that high degree nodes are at particular risk. Our results suggest that deleting the corresponding vector representation from network embeddings alone is not a sufficient measure from a privacy perspective.

    In future work, we will focus on the scalability of the attack approach, on the transfer to directed, weighted, or attributed networks, and on attacks on networks with multiple nodes removed. Further, we will study which modifications to network embeddings can provide a reliable defense against such attacks.


%% file: sections/07_appendix.tex
\begin{table*}[!b]
\caption{Performance of the proposed attack with different classification algorithms (KNN = \emph{K-nearest neighbour classification}, SVC = \emph{support vector machines}, DT = \emph{ decision trees}, RF = \emph{random forests}, AB = \emph{AdaBoost}, GNB =  \emph{Gaussian naive bayes classifiers}) on four datasets.}
\label{tab:appendix}
\setlength\tabcolsep{4pt}
\small\centering \begin{tabular}{lllllllllllllll}
\toprule
           &          & \multicolumn{6}{c}{Barabasi} & \multicolumn{6}{c}{Facebook} \\
            \cmidrule(lr){3-8} \cmidrule(lr){9-14}
        Embedding Algorithm   &          &      KNN &   SVC &    DT &    RF &    AB &     GNB &      KNN &   SVC &    DT &    RF &    AB &     GNB \\
    \midrule HOPE & avg auc &     0.52 &  0.49 &  0.51 &  0.53 &  0.61 &  0.63 &     0.78 &  0.89 &  0.67 &  0.86 &  0.95 &  0.91 \\
           & precision@10 &      0.0 &   0.0 &  0.12 &  0.08 &  0.03 &  0.34 &     0.74 &   0.0 &  0.58 &   0.7 &  0.73 &   0.6 \\
           & macro f1 &      0.0 &   0.0 &  0.05 &   0.0 &   0.0 &  0.16 &     0.48 &   0.0 &  0.43 &  0.44 &   0.5 &  0.34 \\
           & micro f1 &      0.0 &   0.0 &  0.05 &  0.01 &   0.0 &  0.17 &     0.26 &   0.0 &   0.4 &  0.25 &  0.32 &  0.29 \\
    \midrule LINE & avg auc &     0.64 &  0.55 &  0.57 &  0.69 &  0.81 &  0.83 &     0.58 &  0.48 &  0.55 &  0.64 &  0.73 &   0.7 \\
           & precision@10 &      0.5 &   0.0 &  0.37 &  0.55 &  0.59 &  0.51 &     0.24 &   0.0 &  0.13 &  0.07 &   0.2 &  0.45 \\
           & macro f1 &     0.14 &   0.0 &  0.23 &  0.15 &  0.18 &   0.4 &      0.1 &   0.0 &  0.11 &  0.02 &  0.07 &  0.33 \\
           & micro f1 &     0.22 &   0.0 &  0.29 &  0.23 &  0.25 &  0.47 &     0.01 &   0.0 &  0.04 &   0.0 &  0.01 &   0.2 \\
    \midrule node2vec & avg auc &     0.57 &  0.54 &  0.53 &   0.6 &  0.78 &  0.72 &     0.53 &  0.51 &  0.51 &  0.54 &  0.79 &  0.72 \\
           & precision@10 &      0.2 &   0.0 &  0.22 &   0.2 &  0.43 &  0.36 &     0.06 &   0.0 &  0.15 &  0.04 &  0.16 &  0.41 \\
           & macro f1 &     0.01 &   0.0 &   0.1 &  0.01 &  0.04 &  0.19 &      0.0 &   0.0 &  0.05 &   0.0 &   0.0 &  0.17 \\
           & micro f1 &     0.02 &   0.0 &  0.11 &  0.02 &  0.04 &  0.19 &      0.0 &   0.0 &  0.06 &   0.0 &   0.0 &   0.2 \\
    \midrule SDNE & avg auc &     0.52 &  0.46 &   0.5 &  0.54 &   0.6 &  0.61 &     0.53 &  0.53 &  0.51 &  0.56 &  0.71 &  0.73 \\
           & precision@10 &     0.04 &   0.0 &  0.08 &  0.04 &  0.07 &  0.33 &     0.11 &   0.0 &  0.13 &  0.09 &   0.2 &  0.42 \\
           & macro f1 &      0.0 &   0.0 &  0.04 &   0.0 &   0.0 &  0.16 &      0.0 &   0.0 &  0.04 &   0.0 &  0.01 &  0.22 \\
           & micro f1 &      0.0 &   0.0 &  0.05 &   0.0 &   0.0 &  0.14 &      0.0 &   0.0 &  0.05 &   0.0 &   0.0 &  0.15 \\
\bottomrule
\toprule
           &          & \multicolumn{6}{c}{Hamsterster} & \multicolumn{6}{c}{DBLP} \\
            \cmidrule(lr){3-8} \cmidrule(lr){9-14}
        Embedding Algorithm   &          &         KNN &   SVC &    DT &    RF &    AB &     GNB &   KNN &   SVC &    DT &    RF &    AB &     GNB \\
    \midrule HOPE & avg auc &        0.68 &  0.64 &  0.62 &  0.74 &   0.9 &  0.83 &  0.63 &  0.95 &  0.65 &  0.75 &  0.95 &  0.96 \\
           & precision@10 &        0.42 &   0.0 &  0.41 &  0.59 &  0.48 &  0.39 &  0.22 &   0.0 &  0.29 &   0.2 &  0.23 &  0.16 \\
           & macro f1 &        0.22 &   0.0 &  0.31 &  0.41 &  0.39 &  0.19 &  0.07 &   0.0 &  0.29 &  0.03 &  0.18 &  0.19 \\
           & micro f1 &        0.04 &   0.0 &  0.12 &  0.05 &  0.09 &  0.14 &  0.09 &   0.0 &  0.15 &  0.02 &   0.1 &  0.21 \\
    \midrule LINE & avg auc &        0.55 &   0.6 &  0.52 &  0.58 &  0.75 &  0.69 &  0.51 &  0.54 &  0.51 &  0.54 &  0.66 &  0.74 \\
           & precision@10 &        0.01 &   0.0 &  0.16 &  0.05 &   0.1 &  0.29 &   0.0 &   0.0 &  0.07 &  0.04 &  0.09 &  0.14 \\
           & macro f1 &         0.0 &   0.0 &  0.07 &  0.02 &  0.05 &  0.23 &   0.0 &   0.0 &  0.03 &  0.01 &  0.01 &  0.11 \\
           & micro f1 &         0.0 &   0.0 &  0.05 &   0.0 &   0.0 &  0.24 &   0.0 &   0.0 &  0.03 &  0.01 &  0.01 &  0.12 \\
    \midrule node2vec & avg auc &        0.53 &  0.53 &  0.51 &  0.54 &  0.71 &  0.66 &   0.5 &  0.51 &   0.5 &  0.52 &  0.57 &   0.6 \\
           & precision@10 &         0.0 &   0.0 &  0.13 &  0.04 &  0.11 &  0.27 &  0.03 &   0.0 &  0.04 &  0.04 &  0.07 &  0.08 \\
           & macro f1 &         0.0 &   0.0 &  0.04 &   0.0 &   0.0 &  0.19 &   0.0 &   0.0 &  0.01 &   0.0 &   0.0 &  0.06 \\
           & micro f1 &         0.0 &   0.0 &  0.05 &   0.0 &   0.0 &  0.22 &   0.0 &   0.0 &  0.02 &   0.0 &  0.01 &  0.05 \\
    \midrule SDNE & avg auc &        0.52 &  0.53 &   0.5 &  0.57 &  0.73 &  0.72 &  0.51 &  0.53 &  0.51 &  0.53 &  0.59 &  0.65 \\
           & precision@10 &        0.04 &   0.0 &  0.12 &  0.02 &  0.06 &  0.25 &   0.0 &   0.0 &  0.03 &   0.0 &  0.03 &  0.07 \\
           & macro f1 &         0.0 &   0.0 &  0.03 &   0.0 &  0.01 &  0.18 &   0.0 &   0.0 &  0.02 &   0.0 &  0.06 &  0.08 \\
           & micro f1 &        0.01 &   0.0 &  0.05 &   0.0 &   0.0 &  0.18 &   0.0 &   0.0 &  0.01 &   0.0 &  0.02 &  0.06 \\
\bottomrule
\end{tabular}
\end{table*}

\newpage

\appendix
\section{Appendix}
Table~\ref{tab:appendix} provides detailed results for the performance of the proposed attack with different classification algorithms.